\documentclass[sn-mathphys,Numbered]{sn-jnl}

\usepackage{graphicx}%
\usepackage{multirow}%
\usepackage{amsmath,amssymb,amsfonts}%
\usepackage{amsthm}%
\usepackage{mathrsfs}%
\usepackage[title]{appendix}%
\usepackage{xcolor}%
\usepackage{textcomp}%
\usepackage{manyfoot}%
\usepackage{booktabs}%
\usepackage{algorithm}%
\usepackage{algorithmicx}%
\usepackage{algpseudocode}%
\usepackage{listings}%
\usepackage{subcaption}%
\usepackage{caption}%
\usepackage{tabularx}
\usepackage{float}
\usepackage[T1]{fontenc}

\begin{document}

\title[Article Title]{Enhanced Multi-Object Tracking Using Pose-based Virtual Markers in 3x3 Basketball}  

\author[1]{\fnm{Li} \sur{Yin}}\email{li.yin@g.sp.m.is.nagoya-u.ac.jp}
\author[1]{\fnm{Calvin} \sur{Yeung}}\email{yeung.chikwong@g.sp.m.is.nagoya-u.ac.jp}
\author[1]{\fnm{Qingrui} \sur{Hu}}\email{hu.qingrui@g.sp.m.is.nagoya-u.ac.jp}
\author[2]{\fnm{Jun} \sur{Ichikawa}}\email{ichikawa.jun@shizuoka.ac.jp}
\author[3]{\fnm{Hirotsugu} \sur{Azechi}}\email{hazechi@mail.doshisha.ac.jp}
\author[3]{\fnm{Susumu} \sur{Takahashi}}\email{stakahas@mail.doshisha.ac.jp}

\author*[1,4,5]{\fnm{Keisuke} \sur{Fujii}}\email{fujii@i.nagoya-u.ac.jp}

\affil[1]{\orgdiv{Graduate School of Informatics}, \orgname{Nagoya University}, \orgaddress{\street{Chikusa-ku}, \city{Nagoya}, \state{Aichi}, \country{Japan}}}

\affil[2]{\orgdiv{Faculty of Informatics}, \orgname{Shizuoka University}, \orgaddress{\street{Chuo-ku}, \city{Hamamatsu}, \state{Shizuoka}, \country{Japan}}}

\affil[3]{\orgdiv{Laboratory of Cognitive and Behavioral Neuroscience}, {Graduate School of Brain Science}, \orgname{Doshisha University}, \orgaddress{\street{Miyakodani}, \city{Kyotanabe}, \state{Kyoto}, \country{Japan}}}

\affil[4]{\orgdiv{RIKEN Center for Advanced Intelligence Project}, \orgname{1-5}, \orgaddress{\street{Yamadaoka}, \city{Suita}, \state{Osaka},  \country{Japan}}}

\affil[5]{\orgdiv{PRESTO}, \orgname{Japan Science and Technology Agency}, \orgaddress{\city{Kawaguchi}, \state{Saitama}, \country{Japan}}}

\abstract
    {Multi-object tracking (MOT) is crucial for various multi-agent analyses such as evaluating team sports tactics and player movements and performance.
    While pedestrian tracking has advanced with Tracking-by-Detection MOT, team sports like basketball pose unique challenges. These challenges include players’ unpredictable movements, frequent close interactions, and visual similarities that complicate pose labeling and lead to significant occlusions, frequent ID switches, and high manual annotation costs.
    To address these challenges, we propose a novel pose-based virtual marker (VM) MOT method for team sports, named Sports-vmTracking. This method builds on the vmTracking approach developed for multi-animal tracking with active learning. First, we constructed a 3x3 basketball pose dataset for VMs and applied active learning to enhance model performance in generating VMs. Then, we overlaid the VMs on video to identify players, extract their poses with unique IDs, and convert these into bounding boxes for comparison with automated MOT methods.
    Using our 3x3 basketball dataset, we demonstrated that our VM configuration has been highly effective, and reduced the need for manual corrections and labeling during pose model training while maintaining high accuracy. Our approach achieved an average HOTA score of 72.3\%, over 10 points higher than other state-of-the-art methods without VM, and resulted in 0 ID switches. Beyond improving performance in handling occlusions and minimizing ID switches, our framework could substantially increase the time and cost efficiency compared to traditional manual annotation.}

\maketitle

\section{Introduction}\label{Introduction}
    The objective of multi-object tracking (MOT) is to continuously and accurately track multiple objects in video or image sequences, assigning a unique identifier to each. Early MOT research primarily focused on tracking linear and predictable object movements, such as pedestrians and vehicles in autonomous driving scenarios \cite{zhang2022bytetrack, meinhardt2022trackformer, sun2020transtrack, aharon2022bot}. However, with the growing demand for automated tactical analysis in team sports, recent MOT research has shifted towards more complex challenges within the sports domain \cite{ren2018tracknet, liu2021stam, vats2023player, hu2024basketball, huang2024iterative}. These challenges include frequent occlusions due to player density, especially in sports like basketball. Situations such as players blocking each other’s paths (screens), passing the ball closely between teammates (hand-offs), or competing for the ball after a missed shot (rebounding battles) often create severe occlusion scenarios. Additionally, the appearance similarity among players due to uniforms and the unpredictability of their irregular, non-linear movements further complicate the task. Promising results have been demonstrated on large-scale public sports datasets \cite{cui2023sportsmot, cioppa2022soccernet, teamtrack2023}, showcasing the effectiveness of advanced MOT methods in addressing these challenges.

    Tracking-by-detection remains the most widely adopted approach in MOT due to its efficient integration of object detection and tracking. The typical workflow consists of three key steps:
    (1) Object Detection: In each video frame, an object detector identifies and localizes objects of interest, generating bounding boxes that define their spatial positions within the frame.
    (2) Data Association: A tracking algorithm associates detected objects across consecutive frames, ensuring consistent tracking of each object throughout the sequence. This crucial step employs various techniques, including the Hungarian Algorithm \cite{kuhn1955hungarian}, Kalman Filter \cite{kalman1960new}, and deep learning-based methods such as Re-identification (ReID) \cite{zhou2019omni, wang2020towards, he2021transreid}.
    (3) Trajectory Update: The trajectory of each object is continuously refined using detections from both the current and previous frames, ensuring smooth and consistent object tracking over time.
    This workflow effectively handles the complexities of MOT, maintaining object identities and achieving robust tracking performance.
    
    The issue of association caused by heavy occlusion and appearance similarity has always been one of the key challenges in MOT. In sports scenarios, the complexity of irregular movements, frequent occlusions, and appearance similarities due to team uniforms make the association task significantly more challenging. Recent advances in tracking algorithms have introduced innovations in association to tackle challenges unique to sports scenarios \cite{huang2024iterative, ren2018tracknet, liu2021stam, hu2024basketball}.
    Also in the field of multi-animal MOT, the tracked objects often have very similar appearances, which can easily lead to association failures. A pioneering study used virtual markers (VMs) (i.e., adding "virtual" markers to identify individuals on each image by active learning with pose estimation and manual annotation) called 'vmTracking' to enhance object features, which has proven effective in addressing association challenges caused by appearance similarity and occlusions \cite{Azechi2024.02.07.579241}. However, the effectiveness of the VM method in more complex sports scenarios remains unvalidated, and its potential advantages over automated tracking approaches \cite{huang2024iterative, aharon2022bot, hu2024basketball} are still uncertain. Furthermore, the impact of VM quantity and size on tracking results is also unclear. 
    
    To address these challenges, we introduce Sports-vmTracking, a method based on the vmTracking framework \cite{Azechi2024.02.07.579241}, designed specifically to handle the severe occlusion and visual similarity issues commonly encountered in team sports. To validate the effectiveness of Sports-vmTracking, we constructed a 3x3 basketball dataset because the small number of players combined with frequent, severe occlusions in 3x3 basketball provides an ideal setting to test the method's applicability. Our experimental results show that the proposed method significantly reduces missed and incorrect detections, as well as ID switches, effectively mitigating the effects of heavy occlusion and visual similarity on the 3x3 basketball dataset.
    The contributions of this paper are as follows:
    
        \begin{itemize} 
        \item[$\bullet$]We present Sports-vmTracking, an innovative pose-based VM multi-object tracking method designed specifically for team sports. This marks the first application of the VM method in sports, offering advancements in tracking algorithms for scenarios involving heavy occlusions and frequent identity switches, which are common challenges in sports videos.
        
        \item[$\bullet$]Specifically, this work advances computer vision by enhancing the vmTracking pipeline \cite{Azechi2024.02.07.579241}, enabling the use of pre-annotated datasets for automated active learning without manual corrections. Additionally, the proposed method converts human keypoints into bounding boxes, allowing direct comparisons with automated tracking methods \cite{huang2024iterative, aharon2022bot, hu2024basketball} and improving tracking efficiency. 
        
        \item[$\bullet$]We constructed a specialized 3x3 basketball pose dataset based on \cite{ichikawa2024analysis} with 3,817 consecutive frames and 22902 keypoints significantly surpassing the DeepSportRadar Basketball Instants Dataset \cite{Van_Zandycke_2022} and featuring many severe occlusion scenarios.

        \item[$\bullet$] Our method achieves a 72.3\% HOTA score \cite{luiten2021hota}, outperforming state-of-the-art fully automated multi-object tracking algorithms by over 10 percentage points with 0 ID switches. This approach effectively reduces missed and false detections in occlusion-heavy scenarios and addresses challenges posed by visual similarity.
        \end{itemize}

\section{Related work}\label{Related work}
\subsection{Multi-Object Tracking in Sports}
    Multi-object tracking (MOT) is a fundamental and crucial task in sports analytics. Unlike pedestrian and vehicle tracking in autonomous driving, the application of MOT in sports presents unique challenges, such as frequent occlusions, similar player appearances, and rapid, nonlinear movements. Despite these difficulties, researchers have made substantial contributions to MOT across various sports disciplines.
    
    Recent approaches \cite{cao2023ocsort, zhang2022bytetrack, aharon2022bot, peize2021dance, huang2024iterative, hu2024basketball} primarily follow the tracking-by-detection framework, integrating a re-identification network to generate embedding features for data association. Specifically, Vats et al. \cite{vats2023player} propose an approach that enhances tracking performance in ice hockey by incorporating team classification and player identification techniques. Similarly, Yang et al. \cite{yang2021multi} demonstrate that tracking accuracy in football is significantly improved by jointly localizing both the field and the players.

    In addition, Hu et al. \cite{hu2024basketball} introduce a method designed to address complex multi-object occlusion and long-lost ID issues by incorporating the spatial constraints of basketball and projecting players onto the court plane. Huang et al. \cite{huang2024iterative} further tackle the problem of insufficient bounding box overlap in object detection by extending the IoU calculation, thereby enhancing MOT accuracy in sports scenarios. This study aims to further improve the accuracy of MOT in team sports through the proposed method by addressing challenges such as player occlusion and visual similarity among players.

\subsection{Appearance-based Multi-Object Tracking}
    Appearance-based MOT approaches have become a highly effective solution for maintaining object identities across multiple frames in video sequences, especially in crowded or dynamic environments where objects may occlude one another or share similar appearances. These methods leverage visual features to enable robust tracking, and recent advancements in object ReID models \cite{zhang2021fairmot, wojke2017simple, he2021transreid} and training techniques \cite{peize2021dance} have led to the integration of ReID into many tracking algorithms' association processes. In Tracking-by-Detection frameworks, various approaches \cite{Bewley2016_sort, Wojke2018deep, aharon2022bot} utilize ReID models to extract object-specific features, facilitating the identification and re-association of objects with previously detected instances across frames. This process helps preserve object identities even during movement or transformations.

    In the field of animal behavior analysis, similar challenges arise in multi-animal tracking due to appearance similarities and frequent occlusions. A novel method proposed by Azechi et al. \cite{Azechi2024.02.07.579241} addresses these challenges by introducing VM to differentiate and track individual animals, achieving remarkable success in managing occlusion and appearance similarity issues. As opposed to vmTracking, Sports-vmTracking supports both manually annotated active learning and automated active learning using pre-annotated training datasets. Sports-vmTracking extends vmTracking by supporting both manually annotated and automated active learning using pre-annotated training datasets.  Moreover, it converts human keypoints into bounding boxes, enabling seamless comparisons with automated tracking methods and enhancing tracking performance.
    
\subsection{Multi-Object Tracking via Human Pose Estimation}
    Human pose estimation is a computer vision task that detects human body keypoints (e.g., shoulders, elbows, knees) in images or videos. 2D human pose estimation includes two main approaches: top-down  \cite{toshev2014deeppose, xu2022vitpose, wang2020deep}, which first detects individual persons in an image and then applies pose estimation on each detected person’s region, and bottom-up \cite{lauer2022multi, cao2017realtime, pishchulin2016deepcut}, which directly detects all body keypoints in an image and then associates them to form individual poses for each person. Each method has strengths in handling complex scenes and multi-person estimation.
    
    Multi-object tracking via human pose estimation \cite{wang2020combining, iqbal2017posetrack, raaj2019efficient, insafutdinov2017arttrack} offers advantages over traditional bounding box-based tracking methods in scenarios with occlusion and similar appearances. Bounding box-based tracking methods often struggle to maintain tracking continuity when the target is partially occluded. With pose estimation, even if certain parts are occluded, other keypoints can still be detected and tracked, ensuring better continuity of the target in occluded scenarios. Traditional bounding box tracking \cite{zhang2022bytetrack, aharon2022bot, huang2024iterative}, relies on overall appearance features, making it prone to confusion between similar-looking objects (e.g., athletes in similar uniforms). In the present study, we leverage pose information for multi-object tracking. Pose estimation captures individual keypoints, providing fine-grained details that enhance object differentiation in dense scenes, particularly for individuals with similar appearances. 

\section{Proposed Method}\label{Methods}
   In this study, we introduce the Sports-vmTracking method, a multi-object tracking (MOT) algorithm that leverages pose-based virtual markers (VMs) to improve tracking accuracy and performance. To address the challenges of player association in team sports, particularly due to appearance similarities. 
   
   Firstly, Step 1 outlines the virtual marker creation process, covered in Section \ref{subsec: Creation}. In this phase, a multi-agent pose estimation module is trained through active learning to generate VMs, which are subsequently incorporated into the videos. These VMs serve a dual purpose: they distinguish individual data labels in the training dataset for Step 2, and they assign virtual features to players in test videos for Step 3.

    In Step 2, we detail the training process of the single-agent pose estimation module as described in Section \ref{subsec: Training}. The VM-labeled video dataset generated in Step 1 is utilized to train the single-agent pose estimation module through active learning. 

    Finally, Step 3 describes the automated MOT to output bounding boxes virtual marker tracking process, introduced in Section \ref{subsec:Tracking}. The single-agent pose model trained in Step 2 is deployed in the VM-labeled test videos to generate keypoints for each player, identified by a unique ID. These keypoints are subsequently transformed into bounding boxes, which represent the final tracking results. 

\begin{figure}[htbp]
    \centering
    \includegraphics[width=1.0\textwidth]{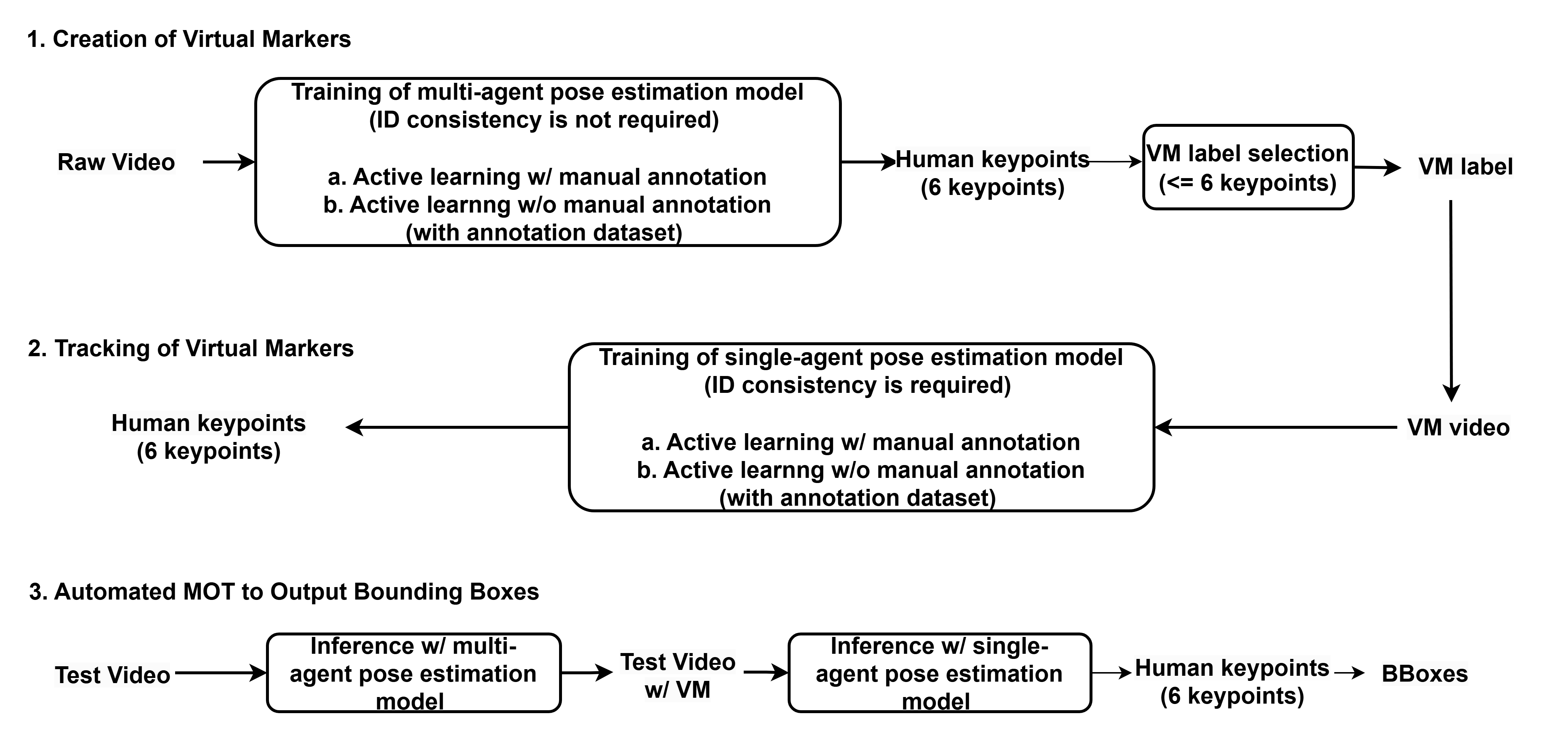}
    
    \caption{The pipeline of Sports-vmTracking. Step 1: Virtual marker creation step. Raw video is used as training data for a multi-agent pose estimation model, with annotations for 6 human keypoints and no requirement for ID consistency between frames. Active learning can further enhance the efficiency of the annotation process. Step 2: Virtual marker tracking step. Similar to Step 1, A single-agent pose estimation model uses VM video data, leveraging VMs as cues to ensure the ID consistency in annotations and predicting 6 keypoints for each player. Step 3:  Automated MOT to Output Bounding Box step. The single-agent pose estimation model is used to infer a test set with VMs, converting the output of 6 human keypoints into bounding boxes to serve as the results for multi-object tracking (MOT).}
    \label{fig: pipeline}
    \end{figure}

    The  Sports-vmTracking method is structured around three main steps, as depicted in Fig \ref{fig: pipeline}.
   

\subsection{Creation of Virtual Markers}\label{subsec: Creation}
    The VM creation process proceeds as follows: Initially, we use DeepLabCut (DLC) \cite{Mathis2018}, an open-source pose estimation toolbox, and select the DLCRNet \cite{lauer2022multi} from the multi-animal DLC mode as the multi-agent pose estimation model, which is known for its efficient training with a small amount of labeled data, resulting in high accuracy in multi-animal pose estimation tasks. The training dataset in this paper is generated by extracting frames from raw video and annotating keypoints for each player using an active learning approach with manual human keypoint annotation. Alternatively, pre-annotated datasets can be utilized for automated active learning without requiring manual correction.
    During the manual annotation process for training the multi-agent pose estimation model, we labeled 6 human keypoints: the head, left and right elbows, left and right ankles, and the center of the hip. To improve robustness, the elbows were labeled instead of the hands, as the rapid movement of the hands often creates detection challenges.
    The model outputs 6 human keypoints (head, left and right elbows, left and right ankles, and the center of the hip) for each player, with different colors used to distinguish between players. At this stage, ID consistency between frames is not required. After manually correcting some ID switches and ensuring the correct colors are assigned to distinguish individuals, all 6 or a subset of the human keypoints will be selected as VMs, this process is referred to as VM label as shown in Fig \ref{fig: pipeline}. These VMs are then overlaid onto the raw video to generate a VM video. The number of VMs per player can be determined to balance workload and effectiveness, which is examined in the experiments section. 
    
    Distinct from the approach used in vmTracking \cite{Azechi2024.02.07.579241}, we divided the data into training and test sets rather than directly processing the videos to be analyzed. In team sports, videos from fixed-position cameras on the field are analyzed to gather data, allowing the trained model to be further used for tracking tasks in the same setting. Additionally, to ensure accurate predictions in highly occluded scenes, we adopted a different approach during the manual annotation phase. Unlike vmTracking, which employs DLC's default k-means \cite{macqueen1967some} clustering method to select key frames for annotation, we prioritized frames with severe occlusions for more targeted annotation. 

\subsection{Tracking of Virtual Markers }\label{subsec: Training}
   The VM tracking process is nearly identical to the VM creation process, with two key differences: (1) To achieve highly precise virtual marker tracking, we selected the single-animal DLC \cite{Mathis2018} project. EfficientNet\_b0 \cite{tan2019efficientnet} model was chosen as the single-agent pose estimation model for its superior performance and lower computational cost compared to other available models, such as ResNet \cite{he2016deep} and MobileNet \cite{howard2017mobilenets}. (2) The training data consists of VM videos, with ID-consistent labeling between frames guided by the VMs. For training the single-agent model, we utilized a pre-annotated training dataset and conducted automated active learning. Compared to the multi-round annotation process in vmTracking \cite{Azechi2024.02.07.579241}, this approach was more convenient and time-efficient.
  
\subsection{Automated MOT to output bounding boxes}\label{subsec:Tracking}
   In this step, we use the trained single-agent pose estimation model to infer the VM test videos. Compared to vmTracking, we added an additional step to convert keypoints into bounding boxes, ensuring that the final MOT results align with standard (automated) MOT output formats. Once the VM test dataset is prepared, the trained single-agent pose estimation model can be used to perform inference on the test data. The resulting human keypoints are then converted into bounding boxes, which facilitate the automated MOT process.

   When converting human keypoints into bounding boxes, we considered two approaches and adopted the second one as the method for this task.
   The first method determines the bounding box based on the maximum and minimum (Max\_Min) coordinates of the keypoints. However, due to the absence of hand keypoints, this approach can result in significant inaccuracies. To mitigate ID switches in this method, the Euclidean distance between corresponding keypoints was calculated, and a threshold was set to exclude points where an ID switch was detected.
   The second method improves on the first by applying an offset (Padding) to the maximum and minimum coordinates, compensating for the missing hand keypoints and reducing inaccuracies. Similarly, this method also calculates the Euclidean distance between corresponding keypoints and sets a threshold to exclude keypoints with detected ID switches. The remaining keypoints are then used to generate bounding boxes, though the results are not accurate.
   
\section{Experiments and Results}\label{Experiments}
    In this section, we detail the experimental setup and results used to validate the effectiveness of Sports-vmTracking for MOT in basketball scenarios. 
    First, in Section \ref{sec: Dataset}, we introduce the dataset, which is divided into pose training and MOT test subsets.
    Second, in Section \ref{sec: Pose}, we describe the pose estimation training process and compare it with vmTracking, highlighting the efficiency and accuracy of our annotation strategies.
    Third, in Section \ref{sec: Benchmark results}, we benchmarked our method against state-of-the-art automated MOT algorithms. The results, evaluated using the HOTA metric \cite{luiten2021hota}, indicate that our Sports-vmTracking approach significantly outperforms these methods.
    Additionally, in Sections \ref{sec: VM Size and Quantity} and \ref{sec: Comparison of Bounding Box}, we analyze the impact of VM size and quantity and also compare two methods for converting keypoints into bounding boxes.
   
\subsection{Dataset}\label{sec: Dataset}
    We constructed a fixed-camera 3x3 basketball video pose dataset \cite{ichikawa2024analysis} consisting of 42 videos with a total of 7,531 frames, and annotated the bounding box data for the 6 players on the court. The dataset includes numerous heavily occluded scenes to validate the effectiveness of our method. We utilized 21 videos, comprising a total of 3,817 frames as the pose training dataset, and another 21 videos, comprising 3,714 frames, as the MOT test dataset. 
    
    Compared to the DeepSportRadar Basketball Instants Dataset \cite{Van_Zandycke_2022}, which contains over 700 images and annotates 4 keypoints per player—the head, hip, and both feet. Our dataset includes enhanced upper limb annotations, adding both elbows keypoints. It comprises 3,817 continuous frames extracted from 21 videos, with 6 labeled keypoints per player: the head, left and right elbows, left and right ankles, and the center of the hip. Additionally, our dataset provides a significantly larger volume of data in the form of continuous video sequences, featuring abundant occlusion scenarios that make it particularly suitable for evaluating tracking performance.

\subsection{Pose Estimation Model Training }\label{sec: Pose}
    DeepLabCut(DLC) 2.2.3 was employed, and all experiments were conducted on a single Titan RTX GPU. DeepLabCut utilizes Root Mean Square Error (RMSE) to assess the discrepancy between model predictions and ground truth values.

    To train the multi-agent pose estimation model, same as vmTracking \cite{Azechi2024.02.07.579241}, the multi-animal project mode of DLC \cite{Mathis2018} was initially selected. The DLC built-in model, DLCRNet \cite{lauer2022multi}, was selected for its high accuracy, robustness, multi-scale feature extraction capabilities, and specialized optimization for multi-animal scenarios. Unlike vmTracking, which annotates 19 keypoints for the human body, we annotated only the minimal required keypoints to reduce workload and enhance efficiency. Specifically, we annotated 6 human keypoints for each player: head, left and right elbows, left and right ankles, and the center of the hip. In the maDLC project’s config.yaml file, the individuals option was configured as player1, player2, ..., player6, and the bodyparts option was set to head, left and right elbows, left and right ankles, and the center of the hip. Initially, a total of 210 frames were extracted for annotation by extracting 10 frames from each video to perform the first round of training, with the number of iterations per round set to 200,000 same as vmTracking. A total of 691 frames were annotated for training, achieving accuracy with a test error of 6.69 pixels, as shown in Table \ref{tab:Pose training}.
    In the training video dataset with VM, we output all 6 keypoints as VMs to facilitate annotation in scenarios with significant occlusion. In the test video dataset with VM, we created 6 test datasets with varying sizes and quantities of VM to evaluate their impact on tracking performance.

    To achieve more accurate VM tracking, the single-animal project mode of DLC (saDLC) was selected. Similar to vmTracking \cite{Azechi2024.02.07.579241}, the DLC built-in EfficientNet\_B0 \cite{tan2019efficientnet} was selected as the single-agent pose estimation model for its superior performance over ResNets \cite{he2016deep}, offering optimal depth, width, and resolution scaling. Within saDLC, the iteration settings, annotated keypoints, and body parts for each player were kept consistent with the previously mentioned settings. Since the config.yaml file in saDLC does not include individual settings, we differentiated each body part by setting bodyparts in the format playerID\_bodypart (e.g., player1\_head, player1\_center, ..., player6\_head, player6\_center, etc.). With 6 players and 6 body parts per player, a total of 36 entries were configured in the bodyparts settings, requiring ID-specific labeling in this step. Even when keypoints are occluded, we annotate them as accurately as possible to improve prediction accuracy in occluded scenarios.
    
    
    
    \begin{table}[h]
    \centering
    \resizebox{\textwidth}{!}{  
    \begin{tabular}{llcc}
    \toprule
    Pose Model          & Annotation Frames & Train RMSE (pixel) & Test RMSE (pixel) \\
    \midrule
    \textbf{Multi-agent pose estimation model}&&&\\
    DLCRnet             & 691/3817              & 4.92                  & 6.69 \\
    \textbf{Single-agent pose estimation model}&&&\\
    EfficientNet\_b0    & 659/3817              & 2.43                  & 4.40  \\
    EfficientNet\_b0    & 3817/3817            & 1.59                  & 4.16 \\
    \bottomrule
    \end{tabular}
    }
    \caption{Training and test errors for the DLCRnet model within maDLC and the EfficientNet\_b0 model within saDLC, trained using active learning. A comparative experiment with full annotation of 3,817 frames was conducted to assess the effectiveness of active learning across EfficientNet\_b0 training.}
    \label{tab:Pose training}
    \end{table}

    To demonstrate the effectiveness of active learning, we conducted a comparative experiment during the training of the EfficientNet\_B0 model within saDLC. In one setup, the entire training dataset of 3,817 frames was annotated, resulting in a test error of 4.16 pixels. In contrast, using an active learning approach, only 659 frames were annotated, achieving a comparable test error of 4.40 pixels, as shown in Table \ref{tab:Pose training}. This result highlights that with active learning, it was possible to annotate only a small portion of the data while maintaining accuracy comparable to annotating the entire dataset.

\subsection{Benchmark Results}\label{sec: Benchmark results} 
    Here, we first describe the performance metric in MOT, and then explain various comparative methods. 
    HOTA (Higher Order Tracking Accuracy) \cite{luiten2021hota} is a recent holistic evaluation metric used in MOT tasks to provide a more comprehensive assessment of tracking algorithm performance. HOTA is designed to address the shortcomings of traditional MOT metrics by balancing detection accuracy and association accuracy, offering a more holistic evaluation standard, especially with greater robustness in target re-identification and complex environments. Compared to other metrics, HOTA places greater emphasis on the overall performance of target detection, localization, and trajectory association. Therefore, we use HOTA as the evaluation metric for MOT. HOTA is composed of DetA (Detection Accuracy), LocA (Localization Accuracy), AssA (Association Accuracy), FP (False Positives), FN (False Negatives), and IDs (ID Switches).
    
    We compare our method with state-of-the-art tracking algorithms. Deep-EIoU \cite{huang2024iterative} achieves competitive performance on two large-scale multi-object sports player tracking datasets, including SportsMOT \cite{cui2023sportsmot} and SoccerNet-Tracking \cite{cioppa2022soccernet}. Basketball-SORT \cite{hu2024basketball} is a tracking algorithm specifically designed for basketball scenarios, demonstrating robust performance and effectiveness in basketball MOT tasks. BOT-SORT \cite{aharon2022bot} combines the strengths of ByteTrack \cite{zhang2022bytetrack} and SORT \cite{Bewley2016_sort}, making it well-suited for complex, occlusion-prone environments and showing outstanding performance in sports video analysis and related applications. We evaluated Deep-EIoU, Basketball-SORT, and BOT-SORT on our test video without VMs, using YOLOv8 \cite{yolov8_ultralytics} as the detector, and compared their results to those of our method. 
    
    Additionally, we incorporated DeepLabCut’s multi-animal project (maDLC) \cite{Mathis2018}, a widely recognized tool for multi-animal tracking, as a baseline to evaluate the impact of VMs. The training conditions mirrored those used in Step 1 of the proposed method for the multi-agent pose estimation model. Specifically, we used the same version of DLC referenced in this study, along with the identical multi-agent pose training dataset and DLCRNet model. The conversion of keypoints into bounding boxes followed the same methodology; however, the evaluations for the maDLC approach were conducted on test videos without the use of VMs. Sports-vmTracking utilizes maDLC to generate VMs, which are then used as data annotation cues in single-agent pose estimation model training and for VM-labeled test datasets.
    The results are presented in Table \ref{tab:compare results} below, and Fig \ref{fig:multiple_images} shows some examples of the results.

    \begin{table}[!htbp]
    \centering
    \parbox{\textwidth}
    {
        \begin{subtable}{\textwidth}
            \centering
            \begin{tabular}{lcccc}
            \toprule
            Method & HOTA & LocA & DetA & AssA \\
            \midrule
            Deep-EIoU \cite{huang2024iterative}       & 58.0 ± 6.0 & 82.1 ± 1.0 & 56.3 ± 4.2 & 60.2 ± 9.6 \\
            BOT-SORT \cite{aharon2022bot}       & 55.7 ± 6.4 & \textbf{82.8 ± 1.0} & 55.0 ± 3.0 & 56.9 ± 11.1 \\
            Basketball-SORT \cite{hu2024basketball}  & 61.5 ± 4.9 & 82.5 ± 1.0 & 58.3 ± 2.9 & 65.1 ± 8.3 \\
            maDLC \cite{lauer2022multi} (w/o VMs)          & 52.4 ± 7.0 & 80.1 ± 1.1 & 58.5 ± 4.5 & 47.6 ± 10.3 \\
            Sports-vmTracking (ours)          & \textbf{72.6 ± 2.7} & 80.1 ± 1.2 & \textbf{69.9 ± 2.9} & \textbf{71.9 ± 3.2} \\
            \bottomrule
            \end{tabular}
        \end{subtable}
    }
        
        \vspace{0.3cm} 
        
    \parbox{\textwidth}{
        \begin{subtable}{\textwidth}
            \centering
            \begin{tabular}{lccc}
            \toprule
            Method & FN & FP & IDs \\
            \midrule
            Deep-EIoU \cite{huang2024iterative}                        & 63.0 ± 34.4 & 277.6 ± 119.5 & 6.2 ± 4.7 \\
            BOT-SORT \cite{aharon2022bot}                         & 58.9 ± 34.1 & 342.8 ± 139.6 & 8.1 ± 5.0 \\
            Basketball-SORT \cite{hu2024basketball}                  & 53.2 ± 24.2 & 231.4 ± 140.9 & 4.3 ± 3.0 \\
            maDLC \cite{lauer2022multi} (w/o VMs)                           & 186.9± 70.4 & 3.6   ± 3.3   & 9.9 ± 4.9 \\
            Sports-vmTracking (ours)          & \textbf{3.0 ± 2.9} & \textbf{3.0 ± 2.9} & \textbf{0.0 ± 0.0} \\
            \bottomrule
            \end{tabular}
        \end{subtable}
    }
    
    \caption{Comparison of different multi-object tracking methods on our 3x3 dataset shows that our method achieves significant improvements across all metrics except for LocA, with particularly notable gains in FN, FP, and ID switches.}
    \label{tab:compare results}
    \end{table}

    \begin{figure}[!htp]
    \centering
    \begin{subfigure}{0.05\textwidth}
        \centering
        \raisebox{0.8cm}{\rotatebox{90}{Before occlusion}}
    \end{subfigure}
    \begin{subfigure}{0.3\textwidth}
        \centering
        \includegraphics[width=\textwidth]{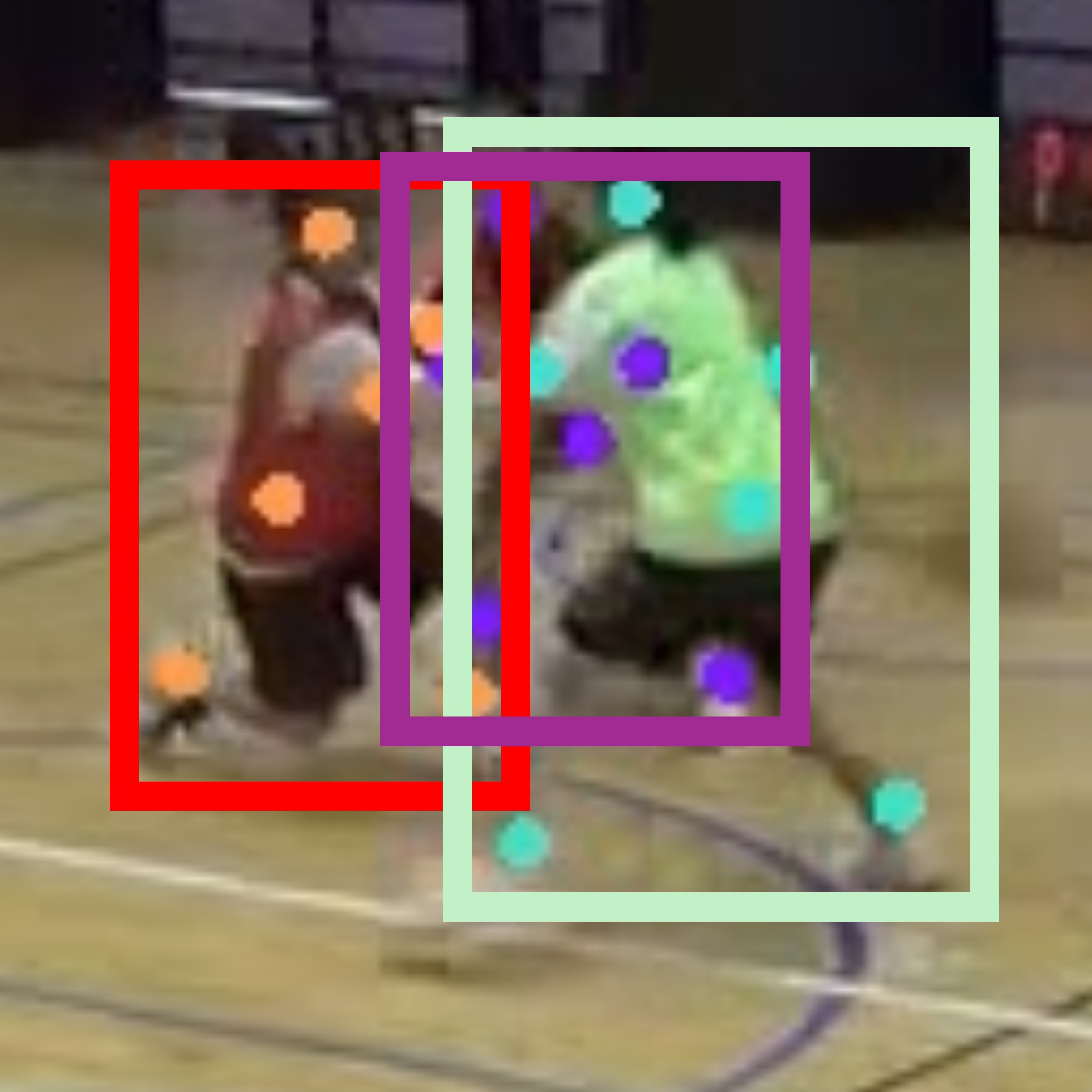}
    \end{subfigure}
    \begin{subfigure}{0.3\textwidth}
        \centering
        \includegraphics[width=\textwidth]{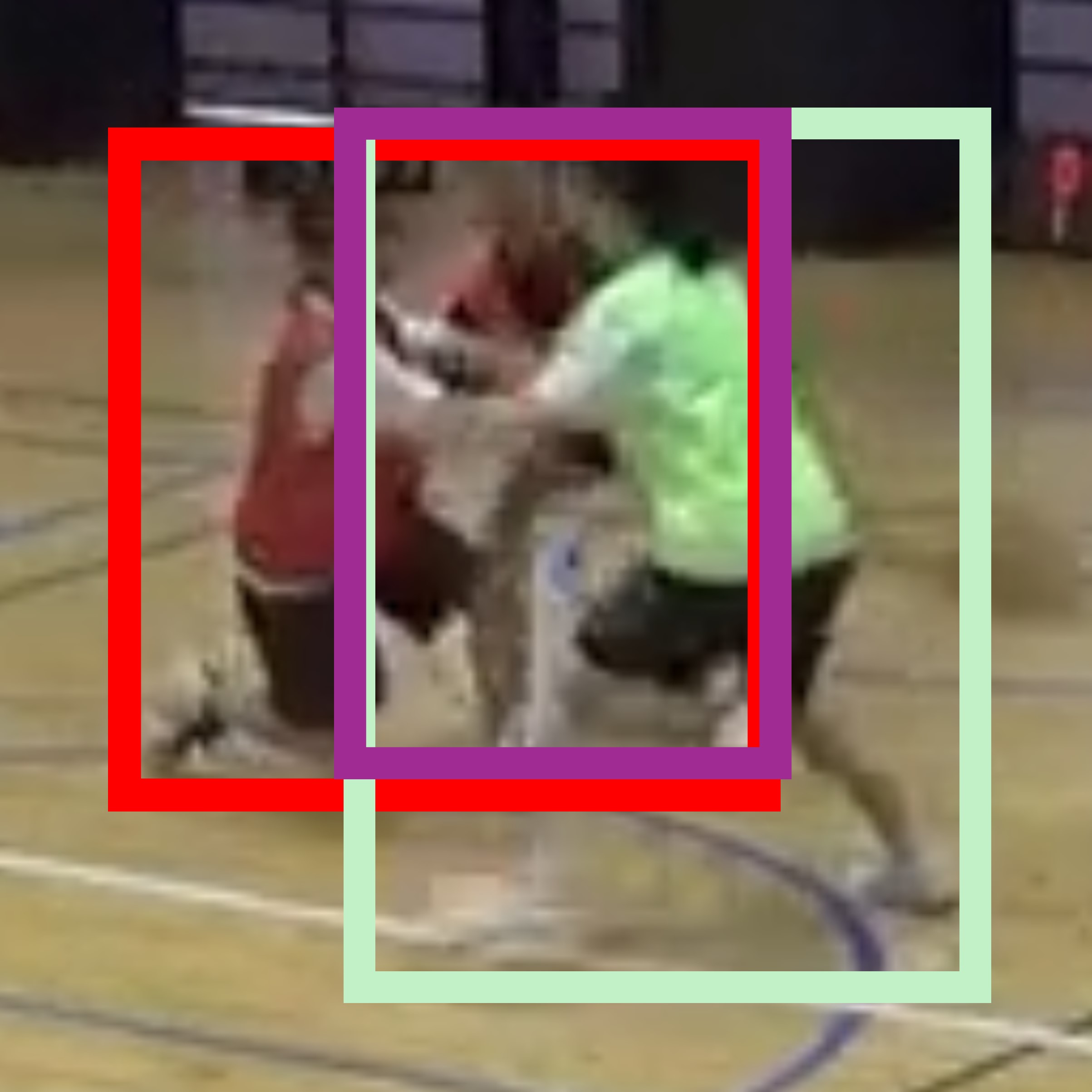}
    \end{subfigure}
    \begin{subfigure}{0.3\textwidth}
        \centering
        \includegraphics[width=\textwidth]{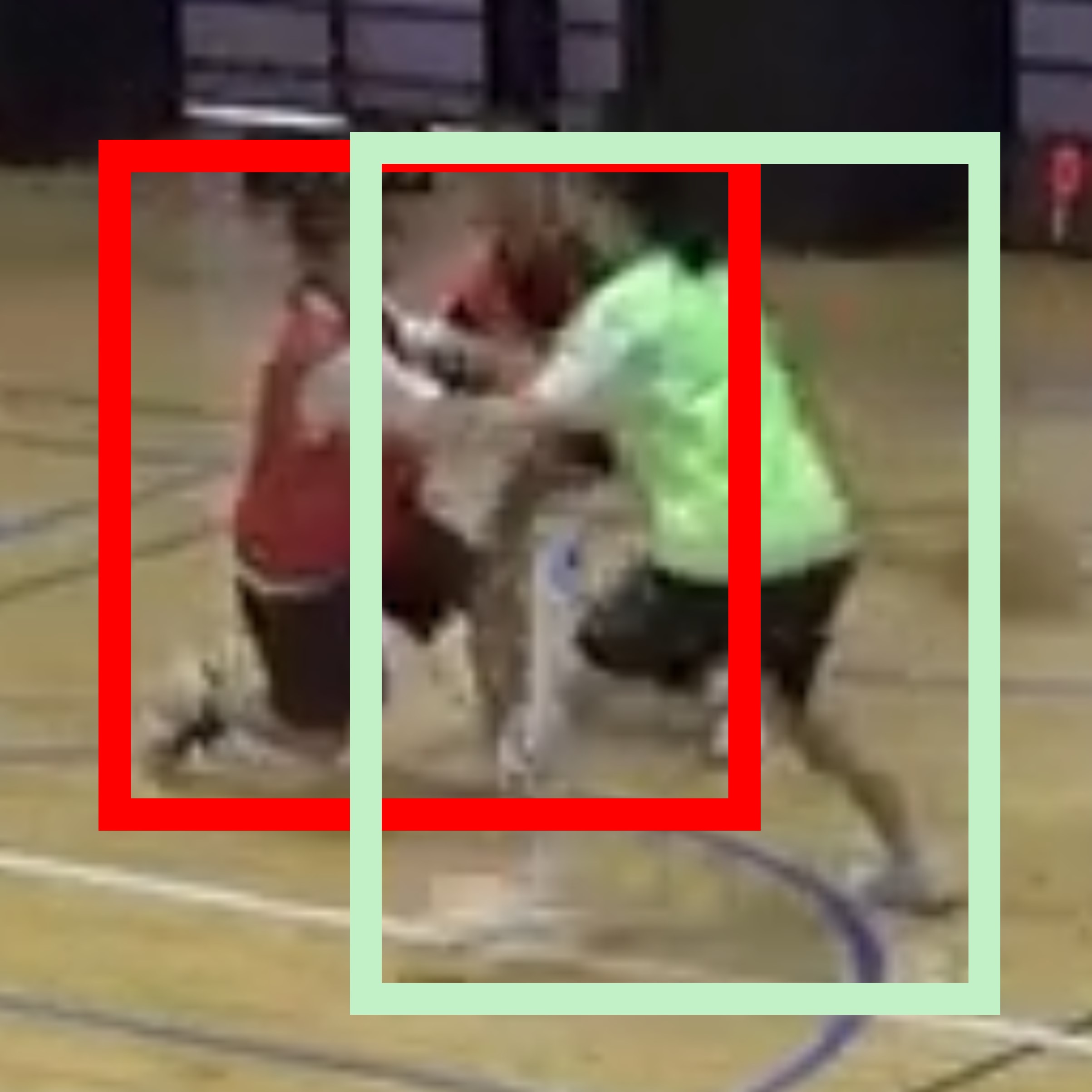}
    \end{subfigure}
    
    \vspace{0.1cm} 
    
    \begin{subfigure}{0.05\textwidth}
        \centering
        \raisebox{1.5cm}{\rotatebox{90}{After occlusion}}
    \end{subfigure}
    \begin{subfigure}{0.3\textwidth}
        \centering
        \includegraphics[width=\textwidth]{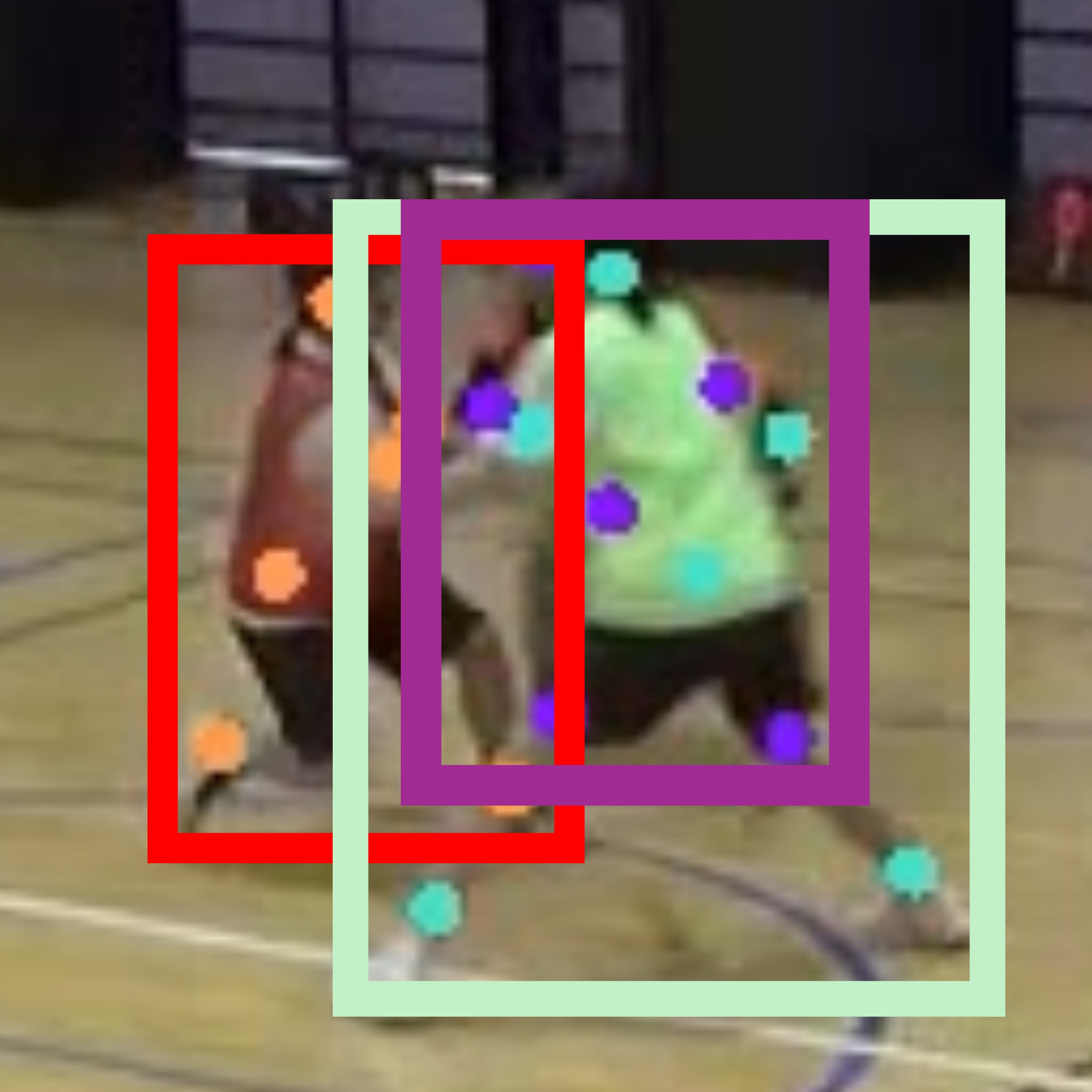}
        \caption{Sports-vmTracking(ours)}
    \end{subfigure}
    \begin{subfigure}{0.3\textwidth}
        \centering
        \includegraphics[width=\textwidth]{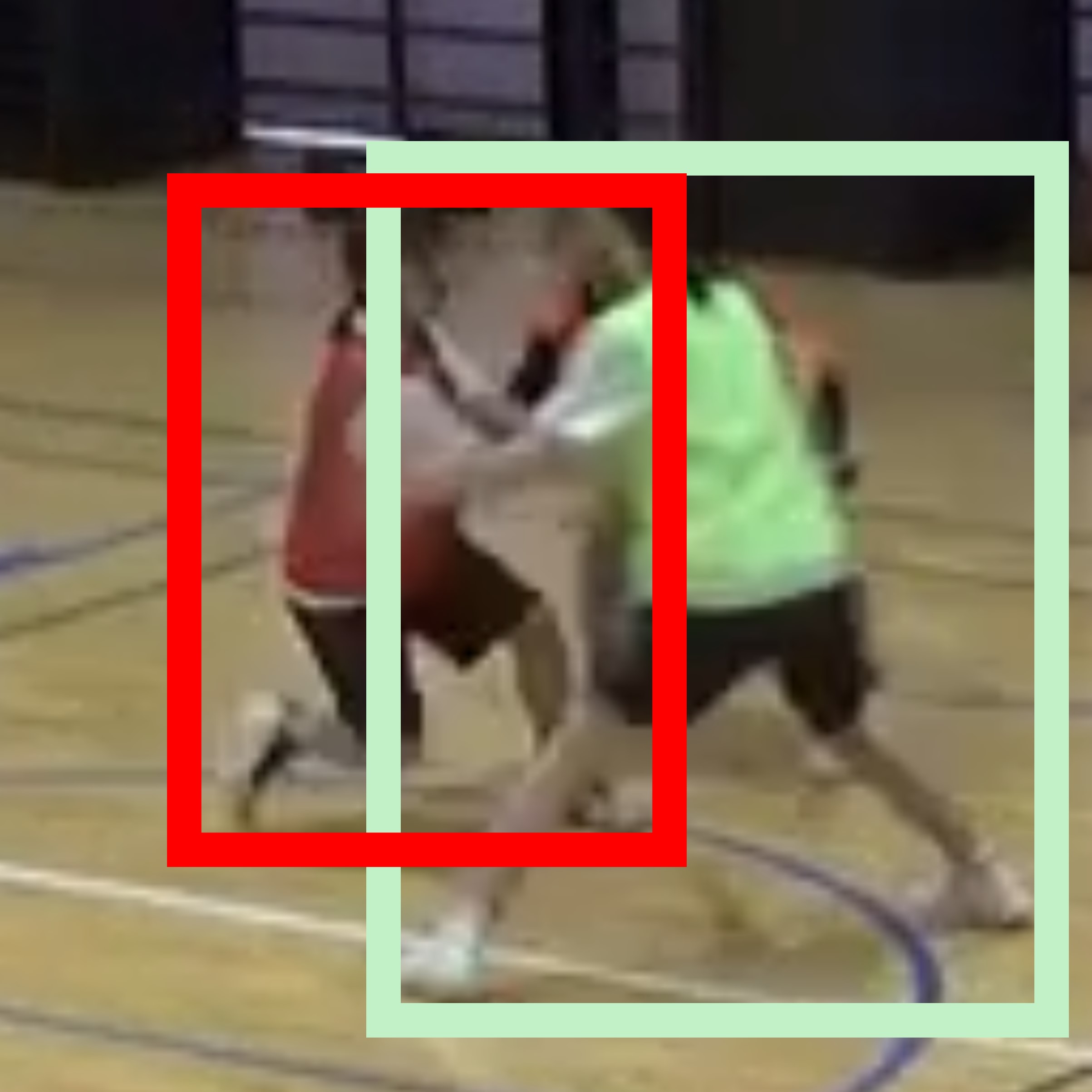}
        \caption{Deep-EIoU\vphantom{Sports-vmTracking (ours)}}
    \end{subfigure}
    \begin{subfigure}{0.3\textwidth}
        \centering
        \includegraphics[width=\textwidth]{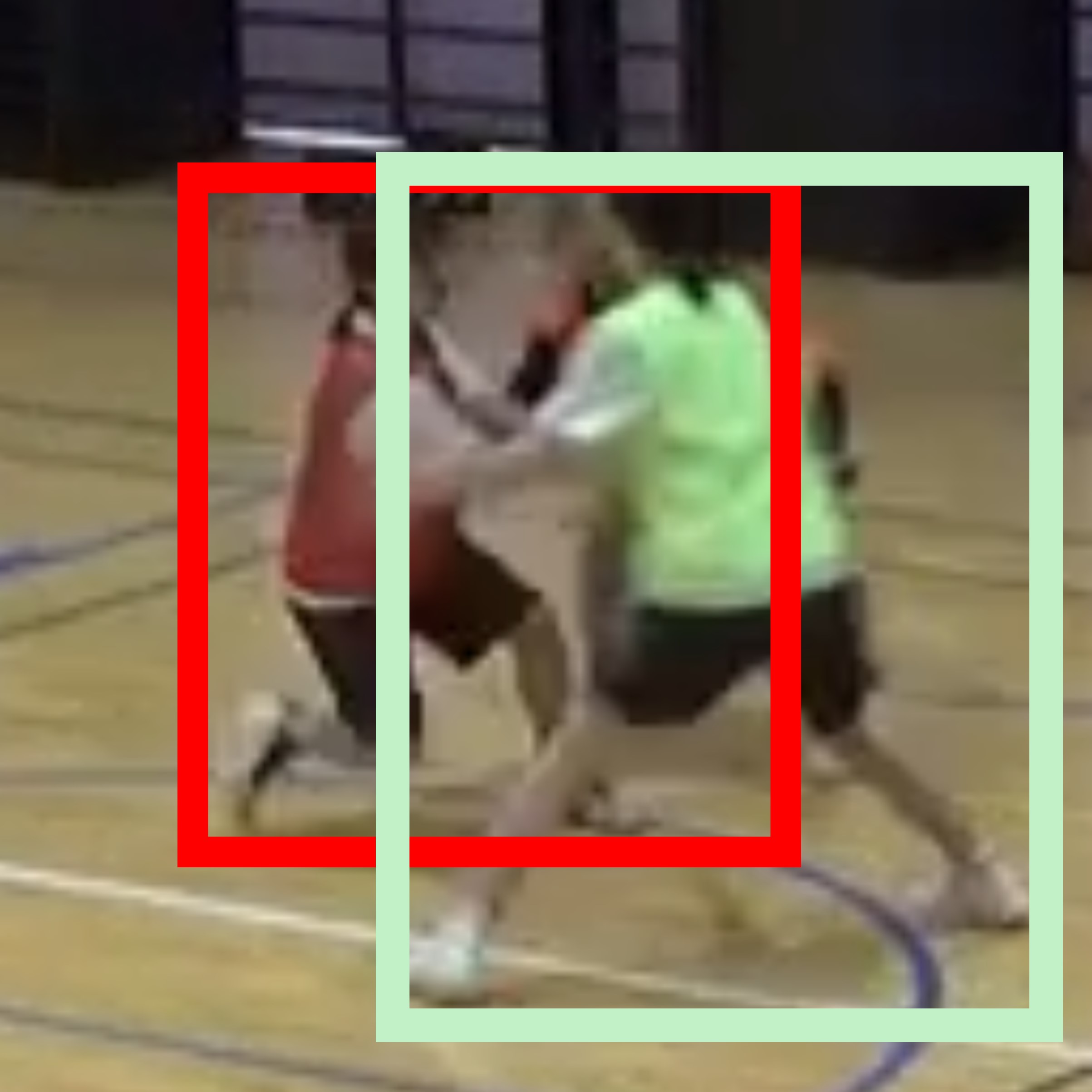}
        \caption{Bot-SORT\vphantom{Sports-vmTracking (ours)}}
    \end{subfigure}
    \caption{The testing results of BOT-SORT and Deep-EIoU on our 3x3 dataset indicate that, under severe occlusion and crowded conditions, Sports-vmTracking more effectively addresses occlusion issues, detecting heavily occluded players (indicated in purple) and maintaining consistent IDs.}
    \label{fig:multiple_images}
\end{figure}

    Sport-vmTracking’s HOTA score exceeds other methods by over 10 points, primarily due to substantial advantages in DetA and AssA. During the active learning process, we annotated a large amount of highly occluded data that the model finds challenging to interpret and predict. As a result, our approach performs well in testing, detecting significantly more heavily occluded scenes with greater accuracy than object detectors. This effectively addresses the object detection and ID switch issues caused by severe occlusions in multi-object tracking. Compared to other methods, our approach yields significantly fewer FP, FN, and ID counts, greatly reducing the complexity of data preprocessing for basketball tactical analysis.
    
\subsection{Impact of VM Size and Quantity} \label{sec: VM Size and Quantity} 
\begin{figure}[!htbp]
    \centering
    \includegraphics[width=0.8\textwidth]{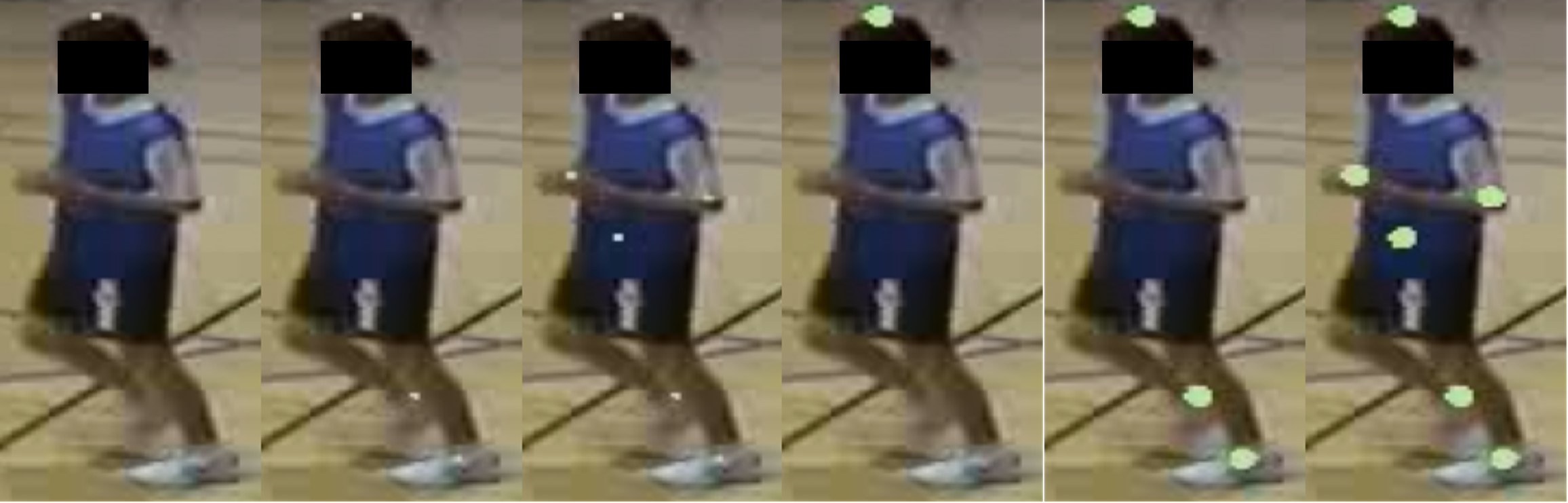}
    \caption{Test video datasets with varying VM sizes and quantities were created. These 6 test sets are used to assess whether differences in VM size and quantity affect the results of pose tracking.}
    \label{fig:markers}
    \end{figure}

    We used the trained EfficientNet\_B0 model to evaluate the test video dataset, yielding the final MOT results. In vmTracking, the VM size was set to a minimum of 1 pixel. To evaluate the impact of varying VM sizes and quantities on tracking performance, we conducted additional experiments, we generated 6 variations of test sets with different VM sizes and quantities for validation. Each player’s VM points were created in two sizes—1 and 3 pixels—across three quantities: 1, 3, or 6 points. The respective keypoint locations are as follows: head (1 point), head and both feet (3 points), and all keypoints—head, both elbows, both feet and center (6 points), as illustrated in Fig \ref{fig:markers}.  The results demonstrate that the smallest VM size (1 pixel) combined with the highest VM quantity (6 points) achieves the best tracking performance.
    
    \begin{table}[!htbp]
    \centering
    \parbox{\textwidth}{
        \begin{subtable}{\textwidth}
        \begin{tabular}{lcccc}
        \toprule
        VM size/num & HOTA & LocA & DetA & AssA \\
        \midrule
        1/1 & 72.5 ± 2.9 & 80.8 ± 1.2 & 71.7 ± 2.8 & 73.4 ± 3.1 \\
        1/3 & 72.4 ± 2.8 & 80.8 ± 1.1 & 71.6 ± 2.7 & 73.3 ± 3.0  \\
        \textbf{1/6} & \textbf{72.6 ± 2.7} & \textbf{80.8 ± 1.1} & \textbf{71.8 ± 2.6} & \textbf{73.5 ± 2.9}\\
        3/1 & 71.8 ± 3.0 & 80.5 ± 1.2 & 71.0 ± 2.9 & 72.7 ± 3.3  \\
        3/3 & 70.8 ± 3.0 & 80.1 ± 1.2 & 69.9 ± 2.9 & 71.9 ± 3.2  \\
        3/6 & 69.6 ± 3.0 & 79.6 ± 1.2 & 68.7 ± 3.0 & 70.7 ± 3.2  \\
        \bottomrule
        \end{tabular}
        \end{subtable}
    }
        \vspace{0.3cm} 
        
   \parbox{\textwidth}{
        \begin{subtable}{\textwidth}
        \begin{tabular}{lcccc}
        \toprule
        VM size/num & FN & FP & IDs \\
        \midrule
        1/1  & 3.1 ± 3.9 & 3.1 ± 3.9 & 0.0 ± 0.0 \\
        1/3  & 2.7 ± 3.1 & 2.7 ± 3.1 & 0.0 ± 0.0 \\
        \textbf{1/6} &  \textbf{2.5 ± 2.9} & \textbf{2.5 ± 2.9} & \textbf{0.0 ± 0.0} \\
        3/1 & 2.6 ± 3.1 & 2.6 ± 3.1 & 0.0 ± 0.0 \\
        3/3 & 3.0 ± 2.9 & 3.0 ± 2.9 & 0.0 ± 0.0 \\
        3/6 & 4.3 ± 4.3 & 4.3 ± 4.3 & 0.5 ± 1.2 \\
        \bottomrule
        \end{tabular}
        \end{subtable}
    }   
    \caption{Experimental results showing the effect of VM size and quantity on tracking performance.}
    \label{tab:vm results}
    \end{table}

     The pose tracking results are summarized in Table \ref{tab:vm results}. As the size of the VM increases, tracking performance generally declines. Specifically, when the VM size was set to 1, it showed the best HOTA performance regardless of the number of VMs, with LocA remaining largely unchanged. However, when the VM size is set to 3, HOTA performance decreases as the number of VMs increases, primarily due to reductions in LocA, DetA, and AssA. Consequently, for VM sizes greater than 1, reducing the number of VMs can improve HOTA performance. The experiments indicate that the optimal results are achieved when the VM size is 1 pixel and the number of VMs is 6.
    
\subsection{Comparison of Bounding Box Generation Methods}  \label{sec: Comparison of Bounding Box} 
    To demonstrate the effectiveness of the padding method in converting human keypoints to bounding boxes, we compared two approaches: (1) using ground truth keypoints with our padding method (Padding). (2) using the method of determining bboxes based on the maximum and minimum (Max\_Min) keypoint values.
    
    The results of the comparison, presented in Table \ref{tab:Bbox}, highlight the significant advantages of the padding method in HOTA metrics. Bounding boxes generated using the padding method are more accurate, achieving higher Intersection over Union (IoU) with the ground truth bounding boxes. This accuracy leads to an increase in true positives (TP), thereby improving LocA, DetA, and AssA scores. Consequently, the overall HOTA performance is enhanced. These findings confirm the effectiveness of the padding method in converting keypoints to bounding boxes, demonstrating its superiority over alternative methods.
    
    \begin{table}[!htbp]
        \centering
        \resizebox{\textwidth}{!}{  
        \begin{tabular}{lccccc}
        \toprule
        Converting method & HOTA & LocA & DetA & AssA & IDs \\
        \midrule
        Padding          & \textbf{75.0 ± 2.0}      & \textbf{81.0 ± 1.0}   & \textbf{74.5 ± 2.1}     & \textbf{75.5 ± 2.1}     & 0.0 ± 0.0 \\
        Max\_Min         & 43.8 ± 2.0      & 66.1 ± 0.7   & 43.2 ± 2.0     & 44.3 ± 2.2     & 0.0 ± 0.0 \\
        \bottomrule
        \end{tabular}
        }
        \caption{Comparison of HOTA scores between the padding method and the max\_min method for converting human keypoints to bounding boxes.}
        \label{tab:Bbox}
    \end{table}

\section{Conclusion}\label{Conclusion}
   In this paper, we present a novel pose-based multi-object tracking method for team sports, termed Sports-vmTracking. We employ an active learning approach to efficiently label training data for generating VMs. These VMs are overlaid on the training data to facilitate individual-specific annotations and are also applied to videos that require tracking, enhancing the visual differentiation of similar players. Our method demonstrates superior performance on our 3x3 basketball multi-object tracking dataset compared to other state-of-the-art tracking algorithms. It effectively tackles challenges such as detection difficulties and ID switches arising from severe occlusions, as well as the complexities of association due to visual similarity, achieving substantial improvements in these areas. 

   While the proposed approach has shown effective performance, certain limitations warrant consideration. For instance, a notable limitation of this study is the scarcity of publicly accessible sports pose datasets. The absence of large-scale and diverse datasets has restricted the validation of our method on widely recognized public benchmarks. Creating sports pose datasets is a time-intensive process that heavily depends on manual annotation. This is due to the complexity of dynamic and occluded movements and the need for precise frame-by-frame labeling, significantly increasing the workload and effort required for dataset construction. Moreover, when annotating the training data for the pose estimation model, we selected keypoints at the head, left and right elbows, left and right ankles, and the center of the hip, omitting hand annotations. This omission could lead to loss of accuracy when converting body keypoints to bounding boxes during prediction. The rapid movement, frequent changes, and smaller size of basketball players’ hands could result in lower prediction accuracy compared to larger body parts like the head or torso.
   In future work, incorporating hand keypoint data and enhancing hand prediction accuracy could further improve the accuracy of multi-player tracking in sports scenes.

\section*{Acknowledgments}
    This work was financially supported by JST SPRING, Grant Number JPMJSP2125, JSPS Grant Number 23H03282, and JST PRESTO Grant Number JPMJPR20CA.
    The author L. Y. would like to take this opportunity to thank the ``THERS Make New
    Standards Program for the Next Generation Researchers''.

\section*{Declarations}
\subsection*{Conflict of Interest}
    The authors declare that they have no conflict of interest.

\subsection*{Compliance with Ethical Standards}
    In the dataset provided from \cite{ichikawa2024analysis}, the participants were fully informed about the study, and their consent was obtained in advance. All the experimental procedures were performed after obtaining approval from the ethical committee at Shizuoka University and Tokoha University. 

\bibliography{sn-bibliography}

\end{document}